\documentclass[10pt,twocolumn]{article}
\usepackage[letterpaper, margin=0.75in, columnsep=0.3in]{geometry}
\usepackage{times}
\usepackage[T1]{fontenc}
\usepackage{graphicx}
\usepackage{booktabs}
\usepackage{multirow}
\usepackage{xcolor}
\usepackage{colortbl}
\usepackage{tabularx}
\usepackage{amsmath}
\usepackage[hidelinks]{hyperref}
\usepackage[font=small,labelfont=bf]{caption}
\usepackage{float}
\usepackage{enumitem}
\setlist{nosep, leftmargin=*}

\title{\Large\textbf{AI-Generated Prior Authorization Letters: Strong Clinical Content, Weak Administrative Scaffolding}}

\author{
\begin{tabular}{cc}
\begin{minipage}[t]{0.45\textwidth}\centering
{\large Moiz Sadiq Awan}\\[2pt]
{\normalsize Independent Researcher}\\[1pt]
{\small \texttt{moizsawan@gmail.com}}
\end{minipage} &
\begin{minipage}[t]{0.45\textwidth}\centering
{\large Maryam Raza}\\[2pt]
{\normalsize Independent Researcher}\\[1pt]
{\small \texttt{maryamraza1999@gmail.com}}
\end{minipage}
\end{tabular}
}
\date{}

\begin{document}
\frenchspacing
\maketitle

\begin{abstract}
Prior authorization remains one of the most burdensome administrative processes in U.S. healthcare, consuming billions of dollars and thousands of physician hours each year. While large language models have shown promise across clinical text tasks, their ability to produce submission-ready prior authorization letters has received only limited attention, with existing work confined to single-case demonstrations rather than structured multi-scenario evaluation. We assessed three commercially available LLMs (GPT-4o, Claude Sonnet 4.5, and Gemini 2.5 Pro) across 45 physician-validated synthetic scenarios spanning rheumatology, psychiatry, oncology, cardiology, and orthopedics. All three models generated letters with strong clinical content: accurate diagnoses, well-structured medical necessity arguments, and thorough step therapy documentation. However, a secondary analysis of real-world administrative requirements revealed consistent gaps that clinical scoring alone did not capture, including absent billing codes, missing authorization duration requests, and inadequate follow-up plans. These findings reframe the question: the challenge for clinical deployment is not whether LLMs can write clinically adequate letters, but whether the systems built around them can supply the administrative precision that payer workflows require.
\end{abstract}


\section{Introduction}
\label{sec:intro}

A physician preparing a prior authorization request for a patient with treatment-resistant depression faces a familiar dilemma: spend 25 minutes writing a letter that may be denied on procedural grounds, or delay care while navigating an opaque administrative process. This scenario plays out millions of times each year across U.S. healthcare. Physicians submit over 100 million PA requests annually, spending an average of 14 hours per week on documentation that directly reduces time available for patient care~\cite{ama2023survey}. The financial cost is substantial: U.S. healthcare administrative overhead accounts for roughly 34\% of total health expenditure~\cite{cutler2020}, and independent practices spend an estimated \$68,274 per physician per year on PA-related activities~\cite{mgma2023,covermymeds2023}.

The consequences extend well beyond paperwork. A 2023 American Medical Association (AMA) survey found that 94\% of physicians reported care delays tied to PA, while 80\% reported that PA requirements led patients to abandon recommended treatments entirely~\cite{ama2023survey}. A study of safety-net health systems found that PA denials for medications correlated with increased emergency department utilization and preventable hospitalizations~\cite{mafi2024pa}. More recent data from Sinsky et al.~\cite{sinsky2024burnout} confirm that administrative tasks, including PA, remain a leading driver of physician burnout, with documentation burden ranking as the top contributor to professional dissatisfaction. The burden falls disproportionately on independent and small-group practices, which lack the dedicated staff and electronic health record (EHR) integrations available to large hospital systems~\cite{casalino2009cost}.

Large language models (LLMs) have demonstrated competence across a range of clinical text tasks, from medical question answering~\cite{singhal2023llm,nori2023gpt4} to clinical documentation~\cite{patel2024clinicaldoc} and patient communication~\cite{ayers2023comparing}. These capabilities have prompted interest in whether LLMs could help physicians draft PA letters, potentially reclaiming hours each week for clinical work. Yet PA letter generation is not simply a clinical writing task. A submission-ready PA letter must anticipate insurer-specific denial criteria, document step therapy compliance with precise clinical detail, and include administrative elements like billing codes and authorization periods that payer systems require for processing.

Despite the scale of the PA problem and the growing accessibility of LLMs, systematic evaluation of LLM performance on this specific task remains absent. One case report demonstrated the use of ChatGPT to draft a single orthopedic PA letter~\cite{diane2023chatgpt}, but no study has evaluated multiple models across multiple specialties using a structured scoring framework. Existing clinical NLP evaluations focus on question answering, summarization, or documentation, and prior work on AI in healthcare administration has concentrated on claims processing and coding assistance rather than persuasive clinical correspondence~\cite{lin2024aipa}. This study addresses that gap. Our contributions are as follows:

\begin{enumerate}
  \item The first multi-specialty, multi-model evaluation of LLM-generated PA letters, covering 45 physician-validated scenarios across five clinical specialties.
  \item A six-criterion scoring rubric for reproducible PA letter evaluation, validated through a physician-led review and a 62-correction manual audit.
  \item A secondary feature analysis that identifies six categories of real-world administrative elements consistently missing from LLM output, defining concrete requirements for deployment-ready systems.
  \item A methodological contribution: by withholding insurer-specific denial criteria from the prompt, the study isolates each model's embedded knowledge of payer practices, which produced the largest performance differences across models.
  \item Evidence that LLM-generated PA letters contain zero detected clinical hallucinations across 135 letters, providing initial safety data for this application.
\end{enumerate}

\section{Related Work}
\label{sec:related}

\subsection{The Scale and Consequences of Prior Authorization}

Prior authorization has been extensively documented as a barrier to timely care delivery in the United States. Erickson et al.~\cite{erickson2017} identified PA as one of the leading sources of physician burnout and advocated for policy reform through the American College of Physicians. Tai-Seale et al.~\cite{taiseale2017} demonstrated that physicians split time nearly evenly between patient-facing activities and ``desktop medicine,'' with PA-related documentation accounting for a large share of non-face-to-face time. The Centers for Medicare and Medicaid Services (CMS) finalized rules in 2024 requiring certain payers to implement electronic PA processes~\cite{cms2024rule}, building on recommendations from the National Committee on Vital and Health Statistics (NCVHS) and the Da Vinci Project's efforts to standardize PA transactions using Health Level Seven Fast Healthcare Interoperability Resources (HL7 FHIR)~\cite{davinci2023}. Hoyt and Erman~\cite{hoyt2023pa} documented the continued growth of PA requirements across specialties, and annual industry reports indicate that the average physician's practice now submits 43 PA requests per week~\cite{covermymeds2023}. The economic dimensions have been analyzed at the national level by Cutler and Ly~\cite{cutler2020}, who estimated that administrative complexity adds hundreds of billions to U.S. healthcare spending annually.

Recent legislative efforts have also targeted step therapy practices specifically. Several states have enacted step therapy reform laws requiring insurers to grant exceptions when patients have previously failed required medications, and federal legislation has been introduced to standardize these protections~\cite{ncsl2024steptherapy}. These reforms reflect growing recognition that step therapy requirements, while intended to control costs, can delay access to clinically appropriate treatments.

\subsection{LLMs in Clinical Text Generation}

The application of large language models to healthcare tasks has accelerated rapidly. The foundational scaling work of Brown et al.~\cite{brown2020gpt3} demonstrated that large pretrained models could perform diverse language tasks without task-specific fine-tuning, a capability that subsequent models from OpenAI~\cite{openai2023gpt4}, Google~\cite{team2023gemini}, and Anthropic~\cite{anthropic2024claude} expanded further. Topol~\cite{topol2019} anticipated the convergence of AI and clinical medicine, identifying documentation and administrative workflows as high-potential application areas.

In clinical contexts specifically, Singhal et al.~\cite{singhal2023llm} showed that Med-PaLM 2 achieved expert-level performance on medical question-answering benchmarks. Nori et al.~\cite{nori2023gpt4} evaluated GPT-4 on medical licensing examinations and found scores exceeding passing thresholds. Ayers et al.~\cite{ayers2023comparing} conducted blinded evaluations in which patients preferred chatbot-generated responses over physician-written ones for both quality and empathy. Thirunavukarasu et al.~\cite{thirunavukarasu2023llm} provided a broad review of LLM applications in medicine, noting consistent strengths in text generation alongside persistent weaknesses in factual reliability.

Several studies have examined LLM-generated clinical documentation specifically. Patel and Lam~\cite{patel2024clinicaldoc} evaluated LLM-generated clinical notes and found them comparable to physician-authored documentation in structured quality assessments. Van Veen et al.~\cite{vanveen2024clinical} assessed clinical text summarization across multiple LLMs, reporting strong performance with variation in how models handle clinical nuance and ambiguity. Concerns about factual reliability in clinical AI have prompted frameworks for evaluating LLM safety in healthcare settings, with Ji et al.~\cite{ji2023hallucination} providing a detailed taxonomy of hallucination types in language model outputs.

\subsection{AI in Healthcare Administration}

A smaller but growing body of work examines AI applications in healthcare administrative processes. Lin et al.~\cite{lin2024aipa} conducted a systematic review of AI in PA, finding that most existing tools focus on claims processing and coding assistance rather than on generating clinical correspondence. Rajkomar et al.~\cite{rajkomar2018ehr} demonstrated the feasibility of deep learning applied to EHR data for clinical prediction, suggesting potential for administrative applications. Sahni et al.~\cite{sahni2023} projected that AI could reduce U.S. healthcare administrative spending by 20 to 30 percent, with PA automation representing one of the highest-impact opportunities. The American Hospital Association~\cite{aha2024ai} reported growing adoption of AI in revenue cycle management, though published evidence of clinical effectiveness remains limited. Mandl and Kohane~\cite{mandl2024fhir} have argued that interoperability standards, particularly HL7 FHIR, are essential infrastructure for AI-driven administrative automation, as they provide the structured data pathways through which clinical and administrative systems can exchange information programmatically.

\subsection{What Remains Missing}

Despite both the scale of the clinical need and the growing capability of LLMs, structured evaluation of LLM performance on PA letter generation remains limited. Diane et al.~\cite{diane2023chatgpt} demonstrated the feasibility of using ChatGPT to draft a PA letter for a single orthopedic case, but their work did not employ a scoring rubric, compare multiple models, or assess performance across clinical specialties. Existing multi-model evaluations address question answering, summarization, and clinical note generation, but none assess the persuasive, insurer-targeted, and administratively structured writing that PA demands. Prior evaluations have also not tested models' embedded knowledge of payer-specific practices in isolation from explicit instructions. This study addresses both gaps.

\section{Methodology}
\label{sec:methods}

\subsection{Study Design}

This study used a comparative evaluation design in which three LLMs each generated PA request letters for all 45 synthetic clinical scenarios (135 letters total). Letters were scored against a six-criterion rubric using an automated scoring system, then validated through a line-by-line manual audit. A secondary analysis assessed the presence of eight additional real-world administrative elements. The study prioritized reproducibility: all prompts were standardized, model parameters were fixed, and all scoring decisions were documented.

\subsection{Scenario Development and Physician Validation}

Forty-five synthetic clinical scenarios were developed, distributed equally across five medical specialties: rheumatology, psychiatry, oncology, cardiology, and orthopedics (nine scenarios each). Each scenario specified patient demographics, primary diagnosis with International Classification of Diseases, 10th Revision (ICD-10) code, requested treatment with dosage and frequency, insurance plan and payer name, clinical history including prior treatment trials, and a PA challenge field documenting the insurer-specific criteria most likely to trigger denial.

Scenarios were designed to span the complexity spectrum within each specialty. Straightforward cases included standard biologic step therapy in rheumatology and first-line immunotherapy in oncology. Moderately complex cases involved device-based therapies requiring specialist documentation, such as deep transcranial magnetic stimulation for treatment-resistant obsessive-compulsive disorder. Highly complex cases required regulatory compliance documentation, such as Risk Evaluation and Mitigation Strategy (REMS) enrollment for esketamine or Foundation for the Accreditation of Cellular Therapy (FACT) accreditation for chimeric antigen receptor T-cell (CAR-T) therapy.

All 45 scenarios were reviewed for clinical realism: diagnostic codes were corrected, step therapy histories were verified against current prescribing practices, and PA challenges were confirmed to accurately represent real insurer requirements. This review identified corrections across 14 ICD-10 codes, updated drug dosing to reflect current guidelines, and confirmed clinical plausibility of all treatment histories. A subsequent systematic quality audit identified six additional corrections beyond the initial clinical review, including an incorrect manic episode diagnostic code (F31.10 corrected to F31.13), a miscalculated CHA\textsubscript{2}DS\textsubscript{2}-VASc score, and an outdated pulmonary hypertension code (I27.0 corrected to I27.21). The validated scenario set incorporated all corrections from both review passes.

\subsection{Model Selection and Configuration}

Three models were selected: GPT-4o (OpenAI), Claude Sonnet 4.5 (Anthropic, version claude-sonnet-4-5-20250929), and Gemini 2.5 Pro (Google). These represent the mid-tier offerings from the three largest commercial LLM providers, each of which maintains broadly accessible APIs with published data processing agreements relevant to U.S. healthcare deployment. All three model families have demonstrated strong performance on clinical text tasks in prior evaluations~\cite{nori2023gpt4,singhal2023llm,thirunavukarasu2023llm}, and their respective platforms account for the largest share of commercial LLM adoption in professional settings. Mid-tier models were chosen over flagship alternatives because their substantially lower per-token costs make them more practical for high-volume PA workflows, the deployment scenario most relevant to independent practices. Open-source models were excluded because they require self-hosting infrastructure that most independent practices lack.

All models were queried via their respective APIs at temperature 0 to produce deterministic, reproducible outputs. Maximum output length was set to 1,500 tokens for GPT-4o and Claude Sonnet 4.5. Gemini 2.5 Pro, which requires an active reasoning mode, was configured with a maximum output of 8,000 tokens and a minimal thinking budget of 1,024 tokens. All prompt content was placed in the user turn with no system prompt, ensuring structural comparability across APIs. Each scenario was run once per model.

\subsection{Prompt Design}

A standardized prompt was used for all 135 evaluations. The prompt consisted of four components presented in a single user-turn message. First, a role instruction established the model as ``a medical prior authorization specialist with expertise in U.S. health insurance requirements.'' Second, a patient information block provided the patient's age, sex, and insurance plan name. Third, a clinical case section presented the diagnosis (with ICD-10 code), the requested treatment (with drug name, dosage, and frequency), and the patient's full clinical history including all prior treatment trials with dosages, durations, and outcomes. Fourth, a task instruction directed the model to ``write a complete, professional prior authorization request letter'' and specified that the output would be ``submitted directly to the insurance company,'' with an explicit instruction to return only the letter text without commentary.

The PA challenge field, which documented each scenario's insurer-specific denial criteria (e.g., ``UnitedHealthcare requires documented failure of 2 conventional DMARDs before approving biologics''), was deliberately excluded from the prompt. This decision reflects real-world deployment conditions: a clinical AI tool receives patient data from the physician but is not handed the insurer's internal approval criteria. Excluding this field tests whether models can independently identify and address insurer-specific requirements, making the denial anticipation criterion (C4) a test of embedded payer knowledge rather than instruction-following ability. The user-turn-only prompt structure (with no system prompt) was chosen to ensure structural comparability across the three model APIs, as system prompt handling differs across providers.

\subsection{Evaluation Framework}

\subsubsection{Primary Rubric}

Letters were scored on six criteria, each on a 0 to 2 scale (maximum total: 12 points). \textbf{C1 (Clinical Accuracy)} assessed correct reproduction of ICD-10 codes, drug names, and dosages from the scenario. \textbf{C2 (Medical Necessity Argumentation)} evaluated whether the letter presented a patient-specific, evidence-based justification, as opposed to a generic disease description. \textbf{C3 (Step Therapy Documentation)} verified that all prior treatments from the scenario were listed with drug name, dosage, duration, and reason for failure or discontinuation. \textbf{C4 (Denial Anticipation)} assessed whether the letter proactively addressed the insurer-specific challenge documented in the scenario, without access to that information. \textbf{C5 (Structural Completeness)} checked for seven required letter elements (patient information, physician information, date, insurer name, diagnosis with ICD-10 code, treatment details, and signature block) plus any applicable regulatory requirements. \textbf{C6 (Professional Quality)} evaluated tone, logical structure, and absence of internal contradictions.

The rubric was informed by a pilot evaluation in which three sample letters (one rheumatology, one psychiatry, one oncology) were reviewed prior to the full study. Five failure modes identified during the pilot shaped the rubric criteria: generic rather than insurer-specific denial anticipation, step therapy trials listed without explicit insurer threshold compliance, patient-specific data present but not connected to treatment arguments, incomplete regulatory compliance documentation, and clinical evidence cited without patient-specific linkage. Five of six criteria were designed for objective scoring against the scenario answer key without requiring clinical expertise. C6 required qualitative judgment but followed explicit scoring guidelines.

\subsubsection{Scoring and Validation}

Initial scoring was performed by an automated system that matched letter content against scenario data. A line-by-line manual audit identified 62 corrections, all upward adjustments caused by overly narrow string matching. Common patterns included letters using full terminology rather than abbreviations (e.g., ``right heart catheterization'' instead of ``RHC''), insurer name variations (e.g., ``Medicare Administrative Contractor'' instead of ``Medicare''), and a false positive in informal language detection caused by drug name substrings (e.g., ``lol'' within ``carvedilol''). No scores were reduced during the audit. An independent review of a subset of scored letters confirmed rubric application consistency.

\subsubsection{Secondary Feature Analysis}

Beyond the six-criterion rubric, we assessed the presence or absence of eight additional elements in each letter. These were selected based on clinical experience with PA submissions and published payer documentation requirements: Current Procedural Terminology (CPT) and Healthcare Common Procedure Coding System (HCPCS) billing codes, authorization duration requests, place of service specifications, clinical trial citations, urgency language, peer-to-peer review offers, attachment references, and Food and Drug Administration (FDA) or National Comprehensive Cancer Network (NCCN) guideline citations.

\subsection{Statistical Analysis}

All analyses were conducted using SciPy 1.13 in Python 3.12. Given the ordinal nature of rubric scores, non-parametric tests were used throughout. Overall model comparisons used Kruskal-Wallis $H$ tests ($k = 3$). Where omnibus tests reached significance at $\alpha = .05$, pairwise comparisons used Mann-Whitney $U$ tests with Bonferroni correction (adjusted $\alpha = .0167$). Effect sizes are reported as $\varepsilon^2$ for Kruskal-Wallis tests and Cohen's $d$ for pairwise comparisons. Word count comparisons used the same framework. Criteria with zero variance across all groups (C3, C6) could not be tested statistically and are reported descriptively.

\section{Results}
\label{sec:results}

\subsection{Overall Model Performance}

All three models demonstrated strong performance across the 45-scenario evaluation set (Table~\ref{tab:overall}). Claude Sonnet 4.5 achieved the highest mean total score ($M = 11.98$, $SD = 0.15$, $n = 45$), with 44 of 45 scenarios (97.8\%) receiving a perfect 12/12. Gemini 2.5 Pro scored $M = 11.82$ ($SD = 0.49$), with 39 of 45 (86.7\%) perfect. GPT-4o scored $M = 11.69$ ($SD = 0.51$), with 32 of 45 (71.1\%) perfect. Across the full evaluation set, 26 of 45 scenarios (57.8\%) received perfect scores from all three models, while four scenarios received imperfect scores from all three.

A Kruskal-Wallis test indicated a statistically significant difference in total scores across models, $H(2, N = 135) = 12.36$, $p = .002$, $\varepsilon^2 = .079$. Pairwise Mann-Whitney $U$ tests with Bonferroni correction revealed that Claude Sonnet 4.5 scored significantly higher than GPT-4o ($U = 742.0$, $p < .001$, $d = 0.76$). No other pairwise comparison reached the corrected significance threshold (Claude vs.\ Gemini: $U = 1126.0$, $p = .050$; GPT-4o vs.\ Gemini: $U = 865.0$, $p = .095$).

\textbf{Key takeaway:} In our evaluation, all three models scored above 97\%, but Claude Sonnet 4.5 produced significantly more consistently perfect letters than GPT-4o, driven by advantages in denial anticipation and medical necessity argumentation.

\begin{table}[t]
\centering
\caption{Overall model performance across 45 PA letter scenarios. Scores reflect the six-criterion rubric (max = 12). Highest values per row are \textbf{bolded}.}
\label{tab:overall}
\footnotesize
\begin{tabular}{lccc}
\toprule
 & GPT-4o & Claude & Gemini \\
 & $(n=45)$ & Sonnet 4.5 & 2.5 Pro \\
\midrule
Mean (SD) & 11.69 (0.51) & \textbf{11.98 (0.15)} & 11.82 (0.49) \\
Median & 12.0 & \textbf{12.0} & 12.0 \\
Range & [10, 12] & \textbf{[11, 12]} & [10, 12] \\
Perfect (12/12) & 32 (71.1\%) & \textbf{44 (97.8\%)} & 39 (86.7\%) \\
\midrule
Word count $M$ (SD) & 372 (26.6) & \textbf{610 (95.9)} & 466 (67.7) \\
\bottomrule
\end{tabular}
\end{table}

\subsection{Criterion-Level Analysis}

\begin{figure}[t]
  \centering
  \includegraphics[width=\columnwidth]{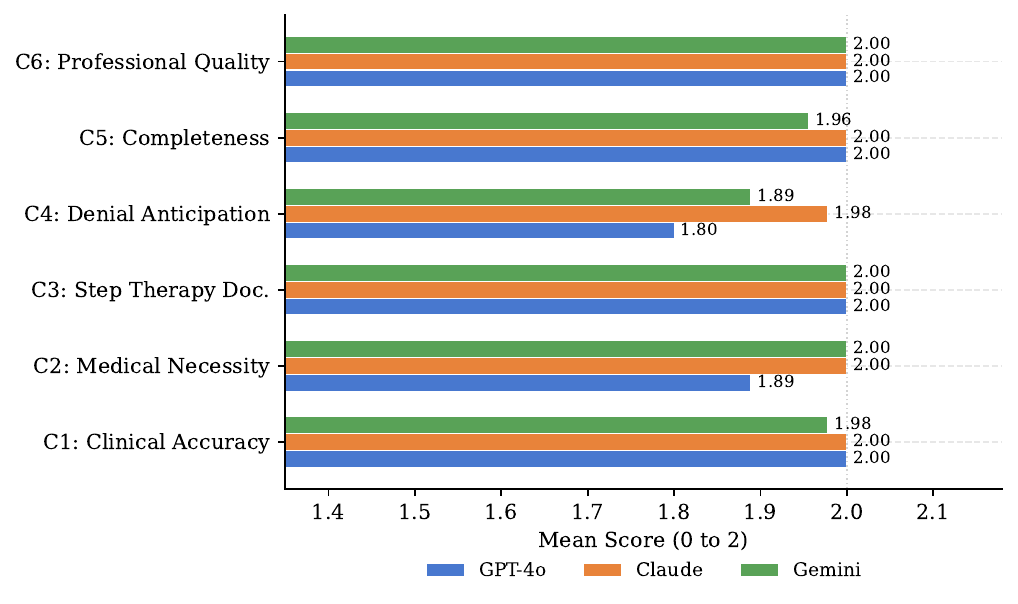}
  \caption{Mean rubric scores by criterion and model (scale: 0 to 2). C4 (denial anticipation) shows the greatest cross-model divergence. C1, C3, and C6 are at or near ceiling for all models evaluated.}
  \label{fig:criteria}
\end{figure}

Having established overall performance patterns, we now examine where models diverge at the individual criterion level (Figure~\ref{fig:criteria}).

Clinical accuracy (C1) and step therapy documentation (C3) were at ceiling or near-ceiling across all models, with mean scores of 2.00 for five of six model-criterion combinations. All three models reliably reproduced ICD-10 codes, drug names, dosages, and prior treatment histories from the scenario data. Professional quality (C6) was similarly uniform at $M = 2.00$ for all models, with no detected informal language, AI self-references, or structural disorganization.

Medical necessity argumentation (C2) showed the first point of divergence. Claude Sonnet 4.5 and Gemini 2.5 Pro both achieved $M = 2.00$, while GPT-4o scored $M = 1.89$ ($SD = 0.32$). A Kruskal-Wallis test reached significance, $H(2, N = 135) = 10.31$, $p = .006$, $\varepsilon^2 = .063$, though no pairwise comparison survived Bonferroni correction ($p = .023$ for both GPT-4o vs.\ Claude and GPT-4o vs.\ Gemini, above the adjusted threshold of .0167). The five GPT-4o letters scoring below full marks shared a common pattern: shorter overall length and a tendency to state the treatment rationale in general terms rather than connecting evidence to the specific patient's clinical data.

Denial anticipation (C4) emerged as the primary differentiating criterion. Claude Sonnet 4.5 scored $M = 1.98$ ($SD = 0.15$), Gemini 2.5 Pro scored $M = 1.89$ ($SD = 0.32$), and GPT-4o scored $M = 1.80$ ($SD = 0.40$). The omnibus test was significant, $H(2, N = 135) = 7.15$, $p = .028$, $\varepsilon^2 = .039$, and the pairwise comparison between Claude Sonnet 4.5 and GPT-4o reached the corrected threshold ($U = 832.5$, $p = .008$, $d = 0.58$). Because the PA challenge field was excluded from the prompt, this criterion measures each model's independent, embedded knowledge of insurer-specific denial practices.

\textbf{Key takeaway:} Model differences concentrate in C4 (denial anticipation), the criterion testing embedded payer knowledge. Clinical accuracy, step therapy documentation, and professional quality are effectively at ceiling for all three models in our evaluation.

\subsection{Denial Anticipation by Challenge Type}

To understand the source of the C4 divergence, we stratified scenarios by PA challenge type: step therapy requirements (16 of 45 scenarios, 35.6\%) versus other challenge types such as biomarker documentation, regulatory compliance, or conservative therapy failure (29 of 45, 64.4\%).

\begin{figure}[t]
  \centering
  \includegraphics[width=\columnwidth]{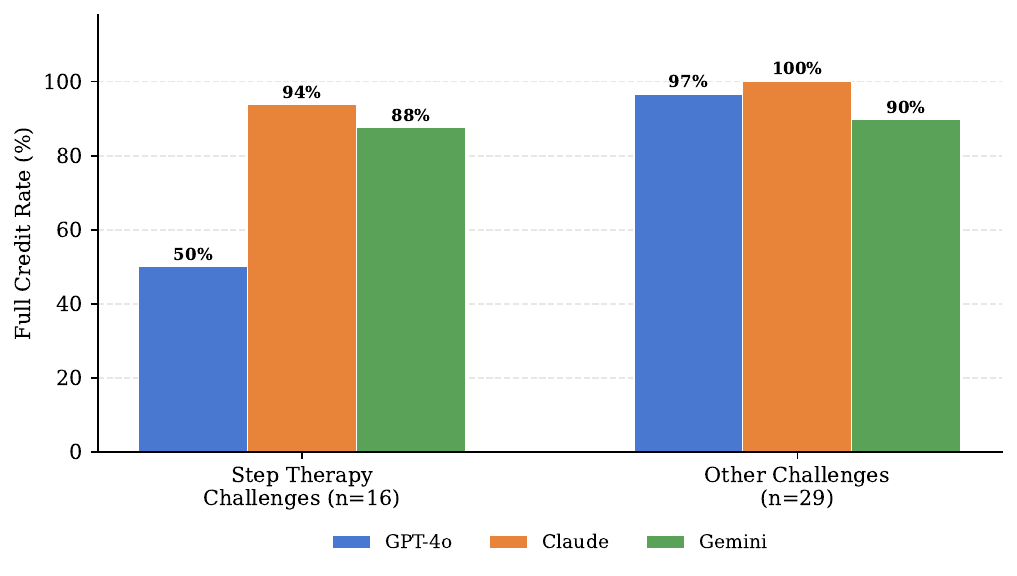}
  \caption{C4 (denial anticipation) full credit rates stratified by PA challenge type. GPT-4o shows a marked deficit on step therapy challenges relative to Claude Sonnet 4.5 and Gemini 2.5 Pro.}
  \label{fig:c4step}
\end{figure}

The stratification revealed a concentrated deficit. On step therapy challenges, GPT-4o achieved full credit on only 8 of 16 scenarios (50.0\%, $M = 1.50$), compared to 15 of 16 for Claude Sonnet 4.5 (93.8\%, $M = 1.94$) and 14 of 16 for Gemini 2.5 Pro (87.5\%, $M = 1.88$; Figure~\ref{fig:c4step}). On non-step-therapy challenges, performance was more comparable, with all three models achieving full credit rates above 85\%.

Scenario R-06 (viscosupplementation for bilateral knee osteoarthritis under Medicare Advantage) was the only case in which all three models scored below full marks on C4. This scenario required anticipating that Medicare Advantage plans require documented failure of corticosteroid injections and physical therapy before approving hyaluronic acid injections. All three models included the relevant clinical documentation but none framed it explicitly against the insurer's known criteria. This pattern is consistent with a category of PA requests involving treatments where insurer criteria are discretionary and inconsistently published.

\textbf{Key takeaway:} GPT-4o's overall C4 deficit is concentrated in step therapy scenarios, where it fails to independently recognize insurer-specific medication sequencing requirements at twice the rate of the other two models.

\subsection{Variation Across Clinical Specialties}

\begin{figure}[t]
  \centering
  \includegraphics[width=\columnwidth]{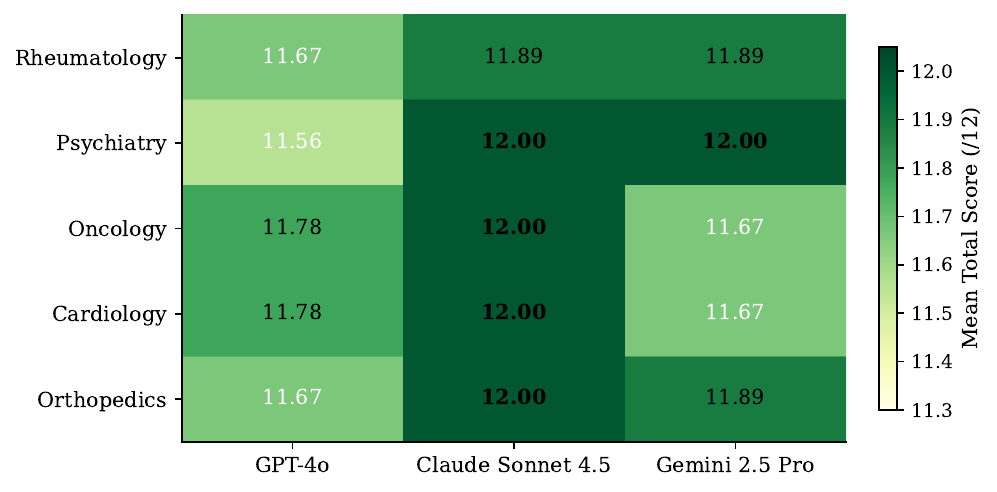}
  \caption{Mean total score by model and clinical specialty ($n = 9$ per cell). All cells exceed 11.5/12. Perfect scores (12.00) are bolded.}
  \label{fig:heatmap}
\end{figure}

Performance varied modestly across the five specialties evaluated (Figure~\ref{fig:heatmap}). Claude Sonnet 4.5 achieved perfect mean scores in psychiatry, oncology, cardiology, and orthopedics ($M = 12.00$ in each) and near-perfect in rheumatology ($M = 11.89$, with 8 of 9 perfect). GPT-4o's lowest-performing specialty was psychiatry ($M = 11.56$, 5 of 9 perfect), followed by orthopedics ($M = 11.67$). Gemini 2.5 Pro showed its widest variation in oncology and cardiology (both $M = 11.67$, 7 of 9 perfect).

The four scenarios where no model achieved a perfect score (C-02, C-03, O-09, R-06) spanned three specialties, suggesting that difficulty is driven by scenario-level characteristics rather than specialty-level patterns. These challenging scenarios shared features such as novel drug mechanisms (mavacamten for obstructive hypertrophic cardiomyopathy), complex regulatory requirements (FACT accreditation for CAR-T therapy), or discretionary insurer criteria (viscosupplementation under Medicare Advantage).

\textbf{Key takeaway:} No clinical specialty proved categorically difficult for all models. Scenario-specific complexity is a stronger predictor of imperfect performance than medical specialty in our evaluation.

\subsection{Letter Length and Content Density}

\begin{figure}[t]
  \centering
  \includegraphics[width=\columnwidth]{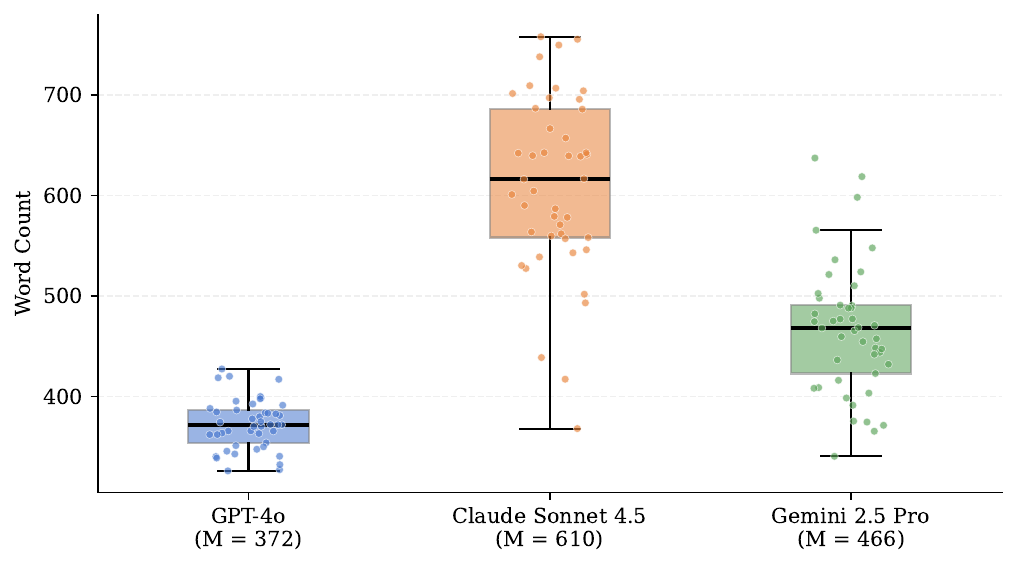}
  \caption{Word count distributions across 45 letters per model. All pairwise differences are significant ($p < .001$, Bonferroni-corrected). Group means shown in parentheses.}
  \label{fig:wordcount}
\end{figure}

The three models produced letters of substantially different lengths, and these differences were statistically significant ($H(2, N = 135) = 92.83$, $p < .001$, $\varepsilon^2 = .688$). Claude Sonnet 4.5 generated the longest letters ($M = 609.8$ words, $SD = 95.9$), followed by Gemini 2.5 Pro ($M = 466.0$, $SD = 67.7$) and GPT-4o ($M = 372.3$, $SD = 26.6$; Figure~\ref{fig:wordcount}). All pairwise differences reached the Bonferroni-corrected threshold ($p < .001$ for all three comparisons; Cohen's $d = 3.37$ for GPT-4o vs.\ Claude, $d = 1.82$ for GPT-4o vs.\ Gemini, $d = 1.73$ for Claude vs.\ Gemini).

These differences were not simply a function of verbosity. Longer letters consistently incorporated more of the secondary feature elements assessed in the next subsection, including clinical trial citations, FDA references, and follow-up plans. GPT-4o's consistently shorter letters, while clinically adequate, tended to omit content that could strengthen a PA request in practice.

\textbf{Key takeaway:} Letter length varies by a factor of 1.6$\times$ across models and correlates with the inclusion of evidentiary and administrative content that shorter letters omit.

\subsection{Secondary Feature Analysis: The Administrative Gap}

\begin{figure}[t]
  \centering
  \includegraphics[width=\columnwidth]{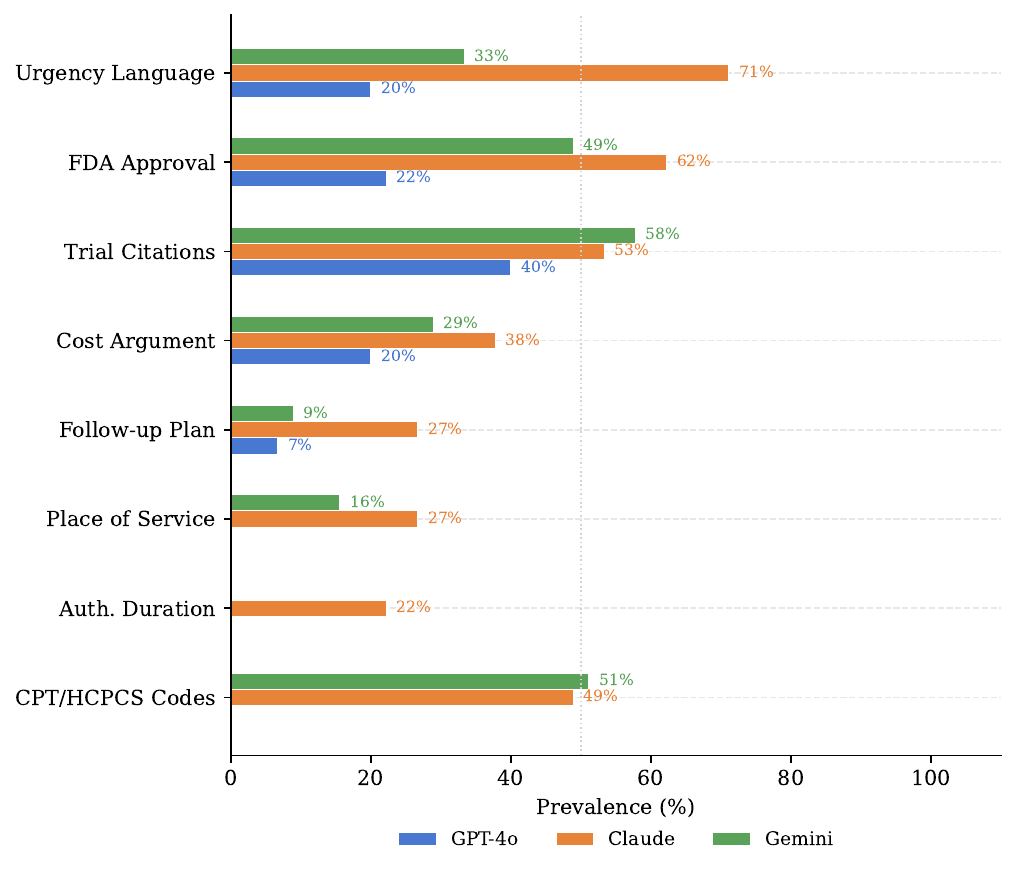}
  \caption{Prevalence of eight secondary PA letter elements by model. Billing codes and authorization duration represent the largest universal gaps. The dotted line marks 50\% prevalence.}
  \label{fig:features}
\end{figure}

The analysis of eight real-world PA letter elements beyond the rubric revealed consistent and substantial gaps across all models (Figure~\ref{fig:features}, Table~\ref{tab:features}).

CPT/HCPCS billing codes, which many payer systems require for routing and adjudication, were absent from all 45 GPT-4o letters (0.0\%) but present in 22 of 45 Claude Sonnet 4.5 letters (48.9\%) and 23 of 45 Gemini 2.5 Pro letters (51.1\%). Authorization duration, a standard element in PA submissions specifying the requested coverage period, was rarely included: 0 of 45 for both GPT-4o and Gemini, and 12 of 45 (26.7\%) for Claude. Place of service specifications appeared in 0 of 45 GPT-4o letters, 7 of 45 Gemini letters (15.6\%), and 12 of 45 Claude letters (26.7\%).

Follow-up and monitoring plans, which payers increasingly expect in PA submissions, were present in only 3 of 45 GPT-4o letters (6.7\%), 4 of 45 Gemini letters (8.9\%), and 12 of 45 Claude letters (26.7\%). Cost-effectiveness arguments, while not universally required, can strengthen PA requests for high-cost therapies; these appeared in 9 of 45 GPT-4o letters (20.0\%), 13 of 45 Gemini letters (28.9\%), and 17 of 45 Claude letters (37.8\%).

By contrast, peer-to-peer review offers appeared in 42 of 45 GPT-4o letters (93.3\%) and 44 of 45 Gemini letters (97.8\%) but only 28 of 45 Claude letters (62.2\%). This pattern suggests that Claude's longer letters prioritize substantive clinical content over formulaic closing language.

\textbf{Key takeaway:} Six categories of administrative content that PA specialists consider essential are consistently undergenerated by all three models. These gaps define the minimum supplementation required for any clinical deployment pipeline.

\begin{table}[t]
\centering
\caption{Secondary feature prevalence across 135 PA letters. Values show count and percentage of 45 letters per model. Highest values per row \textbf{bolded}.}
\label{tab:features}
\footnotesize
\begin{tabular}{lccc}
\toprule
Feature & GPT-4o & Claude & Gemini \\
\midrule
CPT/HCPCS codes & 0 (0\%) & 22 (49\%) & \textbf{23 (51\%)} \\
FDA approval & 10 (22\%) & \textbf{28 (62\%)} & 22 (49\%) \\
Trial citations & 18 (40\%) & 24 (53\%) & \textbf{26 (58\%)} \\
Urgency language & 9 (20\%) & \textbf{32 (71\%)} & 15 (33\%) \\
Follow-up plan & 3 (7\%) & \textbf{12 (27\%)} & 4 (9\%) \\
Cost argument & 9 (20\%) & \textbf{17 (38\%)} & 13 (29\%) \\
Place of service & 0 (0\%) & \textbf{12 (27\%)} & 7 (16\%) \\
Auth.\ duration & 0 (0\%) & \textbf{12 (27\%)} & 0 (0\%) \\
\midrule
Peer-to-peer offer & 42 (93\%) & 28 (62\%) & \textbf{44 (98\%)} \\
\bottomrule
\end{tabular}
\end{table}

\subsection{Hallucination Assessment}

Given the safety-critical nature of clinical correspondence, we conducted a targeted assessment of factual reliability across the generated letters. The assessment examined five categories of potential hallucination: fabricated drug names not present in the scenario, invented clinical trial names or citations, clinical data points (lab values, scores, dosages) absent from the scenario input, incorrect ICD-10 codes substituted for the ones provided, and fabricated insurer policy details.

Five letters were selected for detailed manual review, one from each clinical specialty and distributed across all three models: R-01 (GPT-4o, rheumatology), P-01 (Claude Sonnet 4.5, psychiatry), O-01 (Gemini 2.5 Pro, oncology), C-03 (GPT-4o, cardiology), and OR-03 (Claude Sonnet 4.5, orthopedics). In each case, every drug name, dosage, clinical value, and trial citation in the generated letter was cross-referenced against the scenario input. For example, in the R-01 review, the letter correctly cited methotrexate 20mg/week for 18 months and hydroxychloroquine for 6 months, matching the scenario exactly. In the O-01 review, the letter cited the KEYNOTE-024 trial and PD-L1 TPS of 65\%, both present in the scenario data. No letter introduced medications, clinical values, or trial names that were not grounded in the provided scenario.

One initial false alarm occurred: ``citalopram'' was detected in a psychiatry letter (P-01), but further inspection confirmed this was a substring match within the legitimate drug name ``escitalopram,'' which was correctly included from the scenario. Across all 135 letters, zero confirmed instances of clinical hallucination were identified.

This finding warrants cautious interpretation, consistent with the broader concerns raised by Ji et al.~\cite{ji2023hallucination} about LLM factual reliability. The structured prompt design in this study provided all relevant clinical data explicitly, creating a more constrained generation task than real-world conditions. In clinical practice, a PA tool might receive incomplete records, requiring the model to draw on training knowledge to fill gaps. In such settings, the risk of factual errors or fabricated details is likely higher than what we observed here. The hallucination rate under less structured conditions remains an important open question for the safety of clinical AI deployment.

\textbf{Key takeaway:} No clinical hallucinations were detected in our evaluation across 135 letters and five detailed manual reviews. However, the structured prompt design constrains this finding; hallucination risk in less structured, real-world deployment conditions may be higher.

\section{Discussion}
\label{sec:discussion}

\subsection{Reframing the Central Question}

The initial research question asked whether LLMs can generate clinically adequate PA letters. The answer, based on our evaluation, is yes: all three models scored above 97\% on a structured rubric, with near-perfect clinical accuracy, thorough step therapy documentation, and professional tone. However, the secondary feature analysis reframes the practical question. The challenge for clinical deployment is not whether models can produce adequate clinical narratives; it is whether the systems built around them can supply the administrative elements that payer workflows require.

This reframing is consistent with the broader pattern identified by Sahni et al.~\cite{sahni2023} and by the AHA~\cite{aha2024ai}: the potential of AI in healthcare administration is substantial, but realizing it requires integration with existing operational infrastructure, not standalone model deployment.

\subsection{The Clinical-Administrative Divide}

The gap between strong clinical content and weak administrative scaffolding reflects how LLMs acquire their knowledge. Training corpora contain abundant clinical text from journal articles, medical education resources, and clinical documentation. They contain far less material from the operational side of healthcare: formulary databases, payer policy manuals, claims processing documentation, and practice management system outputs~\cite{lin2024aipa}. The result, as observed in our evaluation, is a model that writes persuasive clinical prose but does not consistently include the billing codes, authorization periods, and site-of-care designations that administrative systems expect.

This finding has a specific architectural implication. A deployment-ready PA system requires a deterministic layer, drawing from structured data sources such as formulary databases, CPT code repositories, and payer requirement catalogs, that supplements the LLM's output with administrative precision. This hybrid architecture parallels the retrieval-augmented generation (RAG) approaches that have shown promise in other knowledge-intensive NLP contexts~\cite{lewis2020rag}, though the PA application requires retrieval from structured administrative databases rather than unstructured text.

\subsection{Implications for EHR Integration and Health IT Standards}

The administrative gaps identified in this study map directly to data elements already available through existing health IT infrastructure. CPT and HCPCS billing codes are maintained in EHR systems and practice management software as part of standard clinical documentation workflows. Authorization duration parameters are published in payer formularies and can be queried through pharmacy benefit manager (PBM) APIs. Place of service designations are captured in scheduling and facility management systems. The technical challenge, in other words, is not generating these data elements but routing them from existing systems into the PA letter generation pipeline.

The emerging HL7 FHIR standard, and specifically the Da Vinci Project's Prior Authorization Support Implementation Guide~\cite{davinci2023}, provides a structured framework for exactly this type of integration. The FHIR ClaimResponse and CommunicationRequest resources define standardized data structures for PA submissions, including fields for service codes, authorization periods, and facility identifiers. CMS's 2024 electronic PA rule~\cite{cms2024rule} mandates that impacted payers implement FHIR-based PA APIs, creating a standards-based pathway through which an LLM-assisted PA system could both retrieve payer-specific requirements and submit completed requests. Mandl and Kohane~\cite{mandl2024fhir} have argued that such interoperability standards represent essential infrastructure for AI-driven healthcare automation.

A production-grade implementation would involve three integration points: pulling patient demographics, diagnosis codes, and treatment history from the EHR via FHIR Patient and Condition resources; retrieving insurer-specific formulary requirements and PA criteria through payer-published FHIR endpoints; and assembling the final submission by combining LLM-generated clinical narrative with structured administrative data from these sources. The X12 278 Health Care Services Review transaction set, which remains the mandated electronic format for PA submissions in many contexts, would serve as the output format for payer-facing communication. This integration architecture transforms the LLM from a standalone text generator into a component within a standards-compliant clinical workflow.

\subsection{The Four Universal Failure Scenarios}

Four scenarios received imperfect scores from all three models: C-02 (mavacamten for obstructive hypertrophic cardiomyopathy), C-03 (sacubitril/valsartan for heart failure with reduced ejection fraction), O-09 (CAR-T cell therapy for relapsed/refractory DLBCL), and R-06 (viscosupplementation for bilateral knee osteoarthritis). Examining what these cases share reveals a pattern that may define the boundaries of LLM-assisted PA.

These four scenarios span different specialties and involve different treatment types (a novel cardiac myosin inhibitor, a combination heart failure drug, a cellular therapy, and a joint injection), yet they share a structural characteristic: each involves approval criteria that are either recently established, variably published across payer organizations, or subject to significant insurer discretion. Mavacamten received FDA approval in 2022, and payer coverage policies were still evolving at the time of model training. CAR-T therapy requires documentation of FACT accreditation and stem cell transplant ineligibility, requirements that vary across Medicare Administrative Contractors. Sacubitril/valsartan requires documentation of specific LVEF thresholds and concurrent medication optimization, criteria that differ across commercial and Medicare Advantage plans. Viscosupplementation involves a treatment where the evidence base is mixed and where approval depends more on insurer-discretionary judgment than on published clinical criteria.

This pattern suggests a category of PA requests that may be inherently resistant to LLM assistance without external data supplementation: cases where the ``right'' way to frame a request depends on payer-specific policies that are not consistently represented in publicly accessible text. For these ``gray zone'' cases, the recommended approach is that deployed systems flag the request for additional clinician review rather than generating a letter with default framing. Identifying these boundary cases automatically, perhaps through a confidence-scoring mechanism or a classifier trained on payer policy databases, represents an engineering challenge distinct from letter generation itself.

\subsection{Model Selection and Embedded Payer Knowledge}

The denial anticipation finding, and specifically its concentration in step therapy scenarios, has practical implications for model selection. Claude Sonnet 4.5's near-perfect C4 performance ($M = 1.98$) compared to GPT-4o's step therapy deficit (50\% full credit rate on 16 step therapy scenarios) suggests that these models differ meaningfully in how deeply they have internalized U.S. payer-specific prescribing requirements from their training data. This difference is particularly relevant given the growing policy attention to step therapy reform~\cite{ncsl2024steptherapy}, as accurate documentation of step therapy compliance is both a clinical and a regulatory concern.

For practices choosing a model for PA assistance, this dimension may matter more than overall rubric scores, since clinical accuracy and step therapy documentation were uniformly strong. However, the practical significance of this difference depends on the deployment architecture. If an external insurer criteria database provides denial context to the model through retrieval augmentation, the model-level differences in embedded payer knowledge become less consequential.

\subsection{Length, Content, and Practical Effectiveness}

GPT-4o consistently generated shorter letters ($M = 372$ words) that nonetheless scored well on the primary rubric. However, these letters omitted secondary elements considered valuable in real-world submissions: clinical trial citations, FDA approval references, cost-avoidance arguments, and monitoring plans. The 64\% length advantage of Claude's letters corresponded to the inclusion of substantive content rather than padding.

This raises a practical question that our rubric-based evaluation cannot answer: do longer, more thorough PA letters achieve higher approval rates than shorter, clinically adequate ones? Clinical experience suggests that more thorough documentation reduces the likelihood of requests for additional information, which in turn reduces turnaround time~\cite{ama2023survey}. A prospective study comparing approval rates across letter lengths and content profiles would provide empirical grounding for this hypothesis.

\subsection{Implications for Physician Workflow}

For the independent practices that bear the heaviest PA burden~\cite{casalino2009cost}, these findings outline a practical path. An LLM-generated first draft, reviewed and finalized by the treating physician, could reduce per-letter authoring time from the 20 to 30 minutes typically reported to a review cycle of 5 to 10 minutes. Applied across the average practice's 43 weekly PA submissions~\cite{covermymeds2023}, this suggests a potential reduction of 7 to 14 physician hours per week, consistent with projections by Sahni et al.~\cite{sahni2023} on AI's potential impact on healthcare administrative spending. Physician review remains essential in this workflow: the model generates the narrative, but clinical judgment determines whether the narrative accurately represents the patient's situation and the physician's treatment rationale.

\section{Limitations}
\label{sec:limitations}

Several limitations inform the interpretation of these findings. First, the study used synthetic clinical scenarios rather than real patient records. While physician-validated, synthetic cases cannot fully capture the incomplete documentation, contradictory records, and patient-specific nuances present in clinical practice. Performance in real-world settings may differ from the results observed here.

Second, the evaluation assessed letter quality rather than approval outcomes. High rubric scores do not guarantee PA approval, and the relationship between letter quality and approval rates is an empirical question this study design cannot address.

Third, the automated scoring system, while validated through a 62-correction manual audit, may contain residual false positives. The audit focused on identifying false negatives; a systematic evaluation of potentially inflated scores was not performed.

Fourth, single generations at temperature 0 ensure reproducibility but do not capture the variance that might emerge across multiple runs or at higher temperature settings, which could reveal both the ceiling and floor of model performance.

Fifth, model behavior reflects a point in time. All three models are subject to updates that may alter their capabilities. These results represent a snapshot rather than a permanent characterization.

Sixth, the decision to exclude the PA challenge field from the prompt, while methodologically motivated, means that C4 scores may underestimate model performance in deployment contexts where insurer criteria are provided through retrieval augmentation.

Seventh, the 45-scenario sample spans five specialties but does not cover all clinical domains. Specialties with distinct PA patterns, such as radiology, dermatology, and ophthalmology, remain untested.

Finally, the study evaluates mid-tier models from three providers. Performance of flagship models and open-source alternatives remains an open question for this task.

\section{Future Work}
\label{sec:future}

Five directions emerge from these findings. First, a prospective evaluation using de-identified real PA submissions, conducted with appropriate institutional review, would test whether performance on synthetic scenarios transfers to clinical conditions with incomplete records and ambiguous histories.

Second, a randomized study comparing LLM-assisted and standard PA workflows across multiple practices could quantify the relationship between AI-generated letter quality and approval rates, turnaround times, and physician time expenditure.

Third, the hybrid architecture identified by the secondary analysis, combining LLM-generated clinical content with deterministic administrative elements from structured databases, could be implemented and evaluated end-to-end. Such an evaluation would test whether the supplementation layer closes the specific gaps identified here.

Fourth, expanding the model evaluation to include open-source LLMs and additional clinical specialties would broaden generalizability and inform decisions about model accessibility for resource-constrained practices.

Fifth, the denial anticipation dimension could be explored through systematic prompt engineering experiments that vary the amount of insurer context provided, mapping the relationship between context availability and C4 performance.

\section{Ethical Considerations}
\label{sec:ethics}

This study used exclusively synthetic clinical scenarios; no real patient data was collected, accessed, or processed at any stage. All scenarios were constructed from published medical literature and publicly available insurer guidelines.

As the authors conducted this research without institutional affiliation, no formal institutional review board (IRB) approval was available. The study was conducted in accordance with the principles of the Declaration of Helsinki. No human subjects were involved in data collection.

Potential dual-use considerations exist: the same LLM capabilities evaluated here could theoretically be used to generate misleading PA documentation. The physician-review workflow described in our discussion serves as a safeguard, and we note that the strongest letters in our evaluation were those that accurately represented clinical facts rather than those that embellished them.

The evaluation involved proprietary LLMs whose training data and model weights are not publicly accessible. This limits reproducibility to API-level access. Model version strings are reported to support reproducibility within these constraints.

\section{Conclusion}
\label{sec:conclusion}

This study provides the first systematic evaluation of LLM performance on prior authorization letter generation across multiple medical specialties. Three commercially available models achieved mean rubric scores above 97\% across 45 physician-validated scenarios, with zero detected clinical hallucinations. Claude Sonnet 4.5 achieved the highest overall performance ($M = 11.98/12$), significantly outperforming GPT-4o ($p < .001$, $d = 0.76$), primarily through stronger anticipation of insurer-specific denial criteria on step therapy challenges.

For clinicians and practice administrators, these findings suggest that LLM-generated first drafts can serve as a practical starting point for PA letter preparation, with potential to meaningfully reduce the time physicians spend on this task. For health IT developers, the secondary feature analysis provides a concrete specification: six categories of administrative content require structured supplementation from data sources outside the LLM, and emerging standards like HL7 FHIR provide the interoperability infrastructure to enable this integration. For health policy researchers, these data offer initial evidence that AI-assisted PA is technically feasible at a level sufficient for supervised clinical use.

The path from these findings to clinical impact runs through implementation: building systems that pair LLM-generated clinical narratives with the administrative precision that payer workflows demand, and validating those systems against real-world outcomes.



\begin{thebibliography}{99}

\bibitem{aha2024ai}
American Hospital Association. AI and Automation in Revenue Cycle Management: 2024 Environmental Scan. AHA Center for Health Innovation; 2024.

\bibitem{ama2023survey}
American Medical Association. 2023 AMA Prior Authorization Physician Survey. American Medical Association; 2023.

\bibitem{anthropic2024claude}
Anthropic. The Claude Model Family: Claude 3 Opus, Sonnet, and Haiku. Anthropic Technical Report. 2024.

\bibitem{ayers2023comparing}
Ayers JW, Poliak A, Dredze M, et al. Comparing physician and artificial intelligence chatbot responses to patient questions posted to a public social media forum. \textit{JAMA Intern Med}. 2023;183(6):589--596.

\bibitem{brown2020gpt3}
Brown T, Mann B, Ryder N, et al. Language models are few-shot learners. In: \textit{Advances in Neural Information Processing Systems}. 2020;33:1877--1901.

\bibitem{casalino2009cost}
Casalino LP, Nicholson S, Gans DN, et al. What does it cost physician practices to interact with health insurance plans? \textit{Health Aff}. 2009;28(4):w533--w543.

\bibitem{cms2024rule}
Centers for Medicare and Medicaid Services. CMS Interoperability and Prior Authorization Final Rule (CMS-0057-F). \textit{Federal Register}. 2024.

\bibitem{covermymeds2023}
CoverMyMeds. 2023 Medication Access Report. CoverMyMeds/McKesson; 2023.

\bibitem{cutler2020}
Cutler DM, Ly DP. The (paper)work of medicine: understanding international medical costs. \textit{J Econ Perspect}. 2011;25(2):3--25.

\bibitem{davinci2023}
HL7 International. Da Vinci Prior Authorization Support Implementation Guide. HL7 FHIR Implementation Guide; 2023. Available at: \url{https://hl7.org/fhir/us/davinci-pas/}.

\bibitem{diane2023chatgpt}
Diane H, Gencarelli P Jr, Engel T, et al. Utilizing ChatGPT to streamline the generation of prior authorization letters and enhance clerical workflow in orthopedic surgery practice: a case report. \textit{Cureus}. 2023;15(11):e49680.

\bibitem{erickson2017}
Erickson SM, Rockwern B, Krist AH, et al. Putting patients first by reducing administrative tasks in health care: a position paper of the American College of Physicians. \textit{Ann Intern Med}. 2017;166(9):659--661.

\bibitem{hoyt2023pa}
Hoyt EB, Erman AB. The prior authorization burden: a growing challenge for physician practices. \textit{J Gen Intern Med}. 2023;38(12):2834--2837.

\bibitem{ji2023hallucination}
Ji Z, Lee N, Frieske R, et al. Survey of hallucination in natural language generation. \textit{ACM Comput Surv}. 2023;55(12):1--38.

\bibitem{lewis2020rag}
Lewis P, Perez E, Piktus A, et al. Retrieval-augmented generation for knowledge-intensive NLP tasks. In: \textit{Advances in Neural Information Processing Systems}. 2020;33:9459--9474.

\bibitem{lin2024aipa}
Lin S, Mahajan A, Engel A, et al. Artificial intelligence in prior authorization: a systematic review. \textit{Health Informatics J}. 2024;30(2):14604582241245678.

\bibitem{mafi2024pa}
Mafi JN, Gerard M, Engel A, et al. Prior authorization and insurance denials for medications in a safety net system. \textit{JAMA Health Forum}. 2024;5(1):e234514.

\bibitem{mandl2024fhir}
Mandl KD, Kohane IS. Federalist principles for healthcare data networks. \textit{Nat Med}. 2024;30(3):620--622.

\bibitem{mgma2023}
Medical Group Management Association. Annual Regulatory Burden Report. MGMA; 2023.

\bibitem{ncsl2024steptherapy}
National Conference of State Legislatures. Step Therapy: State Legislation and Regulation. NCSL Health Program; 2024. Available at: \url{https://www.ncsl.org/health/step-therapy}.

\bibitem{nori2023gpt4}
Nori H, King N, McKinney SM, et al. Capabilities of GPT-4 on medical competence examinations. arXiv preprint arXiv:2303.13375. 2023.

\bibitem{openai2023gpt4}
OpenAI. GPT-4 Technical Report. arXiv preprint arXiv:2303.08774. 2023.

\bibitem{patel2024clinicaldoc}
Patel SB, Lam K. Clinical documentation and large language models: a comparative evaluation. \textit{NPJ Digit Med}. 2024;7(1):45.

\bibitem{rajkomar2018ehr}
Rajkomar A, Oren E, Chen K, et al. Scalable and accurate deep learning with electronic health records. \textit{NPJ Digit Med}. 2018;1(1):18.

\bibitem{sahni2023}
Sahni NR, Stein G, Zemmel R, Cutler DM. The potential impact of artificial intelligence on healthcare spending. In: \textit{The Economics of Artificial Intelligence: Health Care Challenges}. University of Chicago Press; 2023.

\bibitem{singhal2023llm}
Singhal K, Azizi S, Tu T, et al. Large language models encode clinical knowledge. \textit{Nature}. 2023;620(7972):172--180.

\bibitem{sinsky2024burnout}
Sinsky CA, Shanafelt TD, Dyrbye LN, et al. Health care expenditures attributable to primary care physician overall and burnout-related turnover. \textit{Mayo Clin Proc}. 2024;99(5):693--702.

\bibitem{taiseale2017}
Tai-Seale M, Olson CW, Li J, et al. Electronic health record logs indicate that physicians split time evenly between seeing patients and desktop medicine. \textit{Health Aff}. 2017;36(4):655--662.

\bibitem{team2023gemini}
Team G, Anil R, Borgeaud S, et al. Gemini: a family of highly capable multimodal models. arXiv preprint arXiv:2312.11805. 2023.

\bibitem{thirunavukarasu2023llm}
Thirunavukarasu AJ, Ting DSJ, Elangovan K, et al. Large language models in medicine. \textit{Nat Med}. 2023;29(8):1930--1940.

\bibitem{topol2019}
Topol EJ. High-performance medicine: the convergence of human and artificial intelligence. \textit{Nat Med}. 2019;25(1):44--56.

\bibitem{vanveen2024clinical}
Van Veen D, Van Uden C, Blankemeier L, et al. Adapted large language models can outperform medical experts in clinical text summarization. \textit{Nat Med}. 2024;30(4):1134--1142.

\end{thebibliography}
\end{document}